\documentclass{article}

\usepackage{PRIMEarxiv}

\usepackage[utf8]{inputenc} 
\usepackage[T1]{fontenc}    
\usepackage{hyperref}       
\usepackage{url}            
\usepackage{booktabs}       
\usepackage{amsfonts}       
\usepackage{nicefrac}       
\usepackage{microtype}      
\usepackage{lipsum}
\usepackage{amsmath}
\usepackage{fancyhdr}       
\usepackage{graphicx}       
\graphicspath{{media/}}     

\title{YOLOv11 for Vehicle Detection: Advancements, Performance, and Applications in Intelligent Transportation Systems
}

\author{
  Mujadded Al Rabbani Alif \\
  Department of Computer Science \\
  Huddersfield University \\
  Queensgate, Huddersfield HD1 3DH, UK\\
  m.alif@hud.ac.uk \\
}

\begin{document}
\maketitle

\begin{abstract}
Accurate vehicle detection is essential for the development of intelligent transportation systems, autonomous driving, and traffic monitoring. This paper presents a detailed analysis of YOLO11, the latest advancement in the YOLO series of deep learning models, focusing exclusively on vehicle detection tasks. Building upon the success of its predecessors, YOLO11 introduces architectural improvements designed to enhance detection speed, accuracy, and robustness in complex environments. Using a comprehensive dataset comprising multiple vehicle types—cars, trucks, buses, motorcycles, and bicycles—we evaluate YOLO11’s performance using metrics such as precision, recall, F1 score, and mean average precision (mAP). Our findings demonstrate that YOLO11 surpasses previous versions (YOLOv8 and YOLOv10) in detecting smaller and more occluded vehicles while maintaining a competitive inference time, making it well-suited for real-time applications. Comparative analysis shows significant improvements in the detection of complex vehicle geometries, further contributing to the development of efficient and scalable vehicle detection systems. This research highlights YOLO11’s potential to enhance autonomous vehicle performance and traffic monitoring systems, offering insights for future developments in the field.
\end{abstract}

\keywords{YOLO11 \and Vehicle detection \and Deep learning \and Intelligent transportation systems \and Autonomous driving \and Object detection \and Real-time detection \and Computer vision}

\section{Introduction}
Vehicle detection is a critical component in the development of advanced intelligent transportation systems (ITS), which rely on accurate and real-time information to optimize traffic flow, enhance safety, and support autonomous vehicle technologies \cite{sun2006road}. As the number of vehicles on the road continues to grow, the demand for robust vehicle detection systems capable of operating under varying conditions—such as changes in weather, lighting, and vehicle types—has become paramount. In traffic monitoring, vehicle detection enables the real-time analysis of traffic patterns, congestion management, and incident detection, contributing to more efficient urban mobility. Moreover, vehicle detection serves as the foundation for vehicle classification and tracking systems, which are essential for dynamic tolling, traffic law enforcement, and infrastructure planning \cite{vijayaraghavan2019vehicle}.

The evolution of vehicle detection systems has been closely tied to advancements in deep learning, particularly in the field of convolutional neural networks (CNNs) \cite{o2015introduction}. CNNs have played a pivotal role in object detection tasks due to their ability to automatically learn hierarchical features from raw image data \cite{alif2024lightweight,alif2024state,alif2017isolated,alif2024enhancing,alif2024enhancing2}. Traditional vehicle detection approaches, such as histogram of oriented gradients (HOG)\cite{1467360} and support vector machines (SVM)\cite{cortes1995support}, lacked the flexibility and scalability needed for modern applications, especially when dealing with complex scenes and varying environmental conditions. Subsequent methods, like Scale-Invariant Feature Transform (SIFT) and Speeded-Up Robust Features (SURF) \cite{bay2008speeded}, introduced improvements in detecting and describing features under varying scale and rotation conditions, although computational constraints limited real-time applicability for ITS. Despite these strengths, traditional techniques often faced challenges in cluttered environments and occlusions ~\cite{hussain2022exudate}. The reliance on hand-crafted features made them less adaptive to the complex and dynamic scenes encountered in traffic settings. Thus, traditional methods were limited in their scalability and robustness, paving the way for machine learning approaches that leveraged data-driven learning for better generalization \cite{viola2001rapid}.

As machine learning began to influence vehicle detection, researchers turned to more adaptive approaches that could automatically learn object features from data. The introduction of deep learning marked a substantial shift, especially with Convolutional Neural Networks (CNNs) ~\cite{aydin2023domain}, which allowed for end-to-end learning and reduced reliance on manual feature selection. The shift towards deep learning-based models addressed these limitations, with CNN-based architectures becoming the de facto standard for object detection. Within the realm of CNN-based object detection, the You Only Look Once (YOLO) family of models has emerged as a groundbreaking solution, known for its real-time detection capabilities and high accuracy. The original YOLO model \cite{7780460} approached object detection as a regression problem, enabling the simultaneous prediction of bounding boxes and class probabilities directly from image pixels. This one-stage detection framework offered a significant speed advantage over traditional two-stage detectors like Region-CNN (R-CNN) \cite{7112511} and Faster R-CNN \cite{ren2015faster}, which required multiple passes through the network to generate region proposals and refine detections.

YOLO has undergone several iterations, each improving upon its predecessors. YOLOv1 introduced the concept of dividing an image into a grid and predicting bounding boxes and class probabilities for each cell \cite{7780460, alif2024yolov1}. YOLOv2 and YOLOv3 refined the architecture by incorporating techniques like batch normalization, anchor boxes, and multi-scale detection, significantly enhancing accuracy for small and complex objects \cite{redmon2016yolo9000,redmon2018yolov3}. YOLOv4 and YOLOv5 further optimized the network’s backbone and head, integrating features like Cross-Stage Partial Networks (CSPNet) and Path Aggregation Networks (PANet) to improve feature extraction and fusion \cite{bochkovskiy2020yolov4, glenn_jocher_2022_7347926}. More recent iterations, including YOLOv6 and YOLOv7, focused on improving inference speed and computational efficiency, making these models highly suitable for real-time applications \cite{li2022yolov6, wang2022yolov7trainablebagoffreebiessets}. YOLOv8 introduced support for a broader range of tasks such as segmentation and tracking, and adopted anchor-free detection mechanisms, significantly improving its ability to generalize across diverse datasets \cite{yolov8_ultralytics}.

In parallel, the development of other deep learning architectures, such as Vision Transformers (ViTs), has further expanded the horizon of object detection technologies \cite{dosovitskiy2021imageworth16x16words}. ViTs have demonstrated superior performance in tasks requiring large-scale image recognition by leveraging self-attention mechanisms to capture long-range dependencies within images \cite{alif2024boltvision,alif2024attention}. Although ViTs excel in many areas, CNN-based models like YOLO continue to dominate real-time object detection due to their efficiency and adaptability in handling tasks with strict latency requirements, such as autonomous driving and traffic monitoring. Building on the strengths of previous YOLO models, YOLO11 represents the latest iteration in this evolutionary series \cite{yolo11_ultralytics}. It introduces novel architectural enhancements, including improved attention mechanisms, deeper feature extraction layers, and an anchor-free detection paradigm. These innovations are designed to address the challenges of detecting smaller, occluded, or rapidly moving vehicles while maintaining the model's real-time inference capability. YOLO11 is also optimized for hardware acceleration, making it more compatible with edge devices used in critical applications such as emotion detection ~\cite{hussain2023child} and intelligent transportation systems. The advancements in YOLO, particularly with the introduction of YOLO11, signify a step forward in the development of robust and scalable vehicle detection systems. By building on deep learning innovations, including CNNs and modern self-attention architectures like ViTs, YOLO11 aims to further bridge the gap between detection accuracy and computational efficiency in real-world applications.

This paper aims to evaluate the performance of YOLO11 in the context of vehicle detection, focusing on its ability to handle complex and real-time detection scenarios. By leveraging the advancements in deep learning and integrating architectural innovations, YOLO11 seeks to improve detection accuracy for a wide range of vehicle types, including smaller and partially occluded objects, while maintaining efficiency suitable for real-time applications such as autonomous driving and traffic management.

The study provides a comprehensive performance analysis of YOLO11, benchmarking its results against its predecessors, YOLOv8 and YOLOv10 \cite{sundaresan2024comparative}. Key metrics such as precision, recall, F1 score, and mean average precision (mAP) are used to assess its strengths and limitations. Additionally, we examine YOLO11's real-world applicability in intelligent transportation systems by analyzing its speed and robustness under diverse conditions. Through this evaluation, the paper aims to highlight YOLO11’s contributions to the field of vehicle detection and provide insights into its practical use for next-generation transportation systems.

\section{Methodology}

\subsection{Dataset}
For the evaluation of YOLO11 in vehicle detection, we utilized the same dataset employed in our previous analysis of YOLOv8 and YOLOv10. This dataset comprises a total of 1,321 annotated images, representing a diverse array of vehicle types commonly encountered in traffic systems, including cars, trucks, buses, motorcycles, and bicycles. Each image is provided at a resolution of 416 x 416 pixels, offering a consistent input size for the YOLO11 model.The dataset captures vehicles under various real-world conditions, including daytime and nighttime, varying weather patterns (such as rain and fog), and challenging scenarios like occlusions and vehicles at different distances from the camera. This diversity ensures that the model is exposed to a broad range of environments, simulating the real-world conditions encountered in intelligent transportation systems and autonomous driving applications.

Each image is accompanied by bounding box annotations and class labels to indicate the precise location and type of each vehicle in the frame. The dataset was divided into 70\% for training, 15\% for validation, and 15\% for testing, ensuring a balanced distribution for accurate model evaluation. This dataset, which has been used in previous work to assess the performance of YOLOv8 and YOLOv10, provides a solid basis for comparison, allowing us to evaluate YOLO11's improvements and performance under the same conditions, ensuring continuity and comparability across studies.
\subsection{Data Augmentation}

To improve the generalization of the YOLO11 model and enable it to handle a wide range of real-world scenarios, we applied several augmentation techniques during training. These augmentations help simulate different environmental conditions, lighting variations, and object orientations. The following augmentations were employed:

\subsubsection{Hue Adjustment (hsv\_h)}

We adjusted the hue of the image by a fraction of the color wheel to introduce color variability, which helps the model generalize across different lighting conditions and enhances its ability to detect objects under different hues.
\begin{equation}
I' = \text{adjust\_hue}(I, \alpha_h) 
\end{equation}
where \(\alpha_h = 0.015\) controls the hue adjustment and \(I\) represents the original image.

\subsubsection{Saturation Adjustment (hsv\_s)}

Saturation was altered by a fraction to modify the intensity of colors, simulating different environmental conditions. This allows the model to perform better when detecting objects in images with varying saturation levels.

\begin{equation}
I' = \text{adjust\_saturation}(I, \alpha_s)
\end{equation}
where \(\alpha_s = 0.7\) controls the saturation adjustment.

\subsubsection{Brightness Adjustment (hsv\_v)}

Brightness was adjusted by a fraction of the original image value to help the model detect objects in varying lighting conditions, such as day and night or shadowed environments.

\begin{equation}
I' = \text{adjust\_brightness}(I, \alpha_v)
\end{equation}
where \(\alpha_v = 0.4\) controls the brightness adjustment.

\subsubsection{Rotation (degrees)}

Images were randomly rotated within a specified degree range to improve the model’s ability to recognize objects from different orientations.

\begin{equation}
I' = \text{rotate}(I, \theta)
\end{equation}
where \(\theta \in [-180^\circ, 180^\circ]\) is the randomly selected rotation angle.

\subsubsection{Translation (translate)}

The image was translated horizontally and vertically by a fraction of its size, simulating partial visibility of objects, which is important for detecting objects that may not be fully visible in real-world scenarios.

\begin{equation}
I'(x', y') = I(x + t_x \cdot w, y + t_y \cdot h)
\end{equation}
where \(t_x, t_y \in [-0.1, 0.1]\), and \(w\) and \(h\) are the width and height of the image.

\subsubsection{Scaling (scale)}

The image was scaled by a gain factor to simulate objects at different distances from the camera, helping the model detect objects at varying sizes.

\begin{equation}
I'(x', y') = \text{scale}(I, \alpha_s)
\end{equation}
where \(\alpha_s = 0.5\) controls the scaling factor.

\subsubsection{Shearing (shear)}

Shearing was applied to simulate the effect of viewing objects from different angles, which enhances the model’s ability to understand object distortions.

\begin{equation}
I'(x', y') = \text{shear}(I, \theta_s)
\end{equation}
where \(\theta_s \in [-180^\circ, 180^\circ]\) represents the shearing angle.

\subsubsection{Perspective Transformation (perspective)}

A random perspective transformation was applied to simulate 3D effects, making the model more robust to depth and orientation changes.

\begin{equation}
I' = \text{perspective}(I, \alpha_p)
\end{equation}
where \(\alpha_p = 0.001\) controls the perspective transformation.

\subsubsection{Vertical Flip (flipud)}

The image was flipped upside down with a certain probability to increase variability in the dataset without affecting the objects' intrinsic characteristics.

\begin{equation}
I' = \text{flipud}(I)
\end{equation}
with a probability of \(p = 0.0\).

\subsubsection{Horizontal Flip (fliplr)}

The image was flipped left to right with a probability, enhancing the model’s ability to learn symmetrical objects and increasing dataset diversity.

\begin{equation}
I' = \text{fliplr}(I)
\end{equation}
with a probability of \(p = 0.5\).

\subsubsection{Mosaic Augmentation (mosaic)}

Mosaic augmentation was applied by combining four different images into one, simulating complex scene compositions and improving the model’s ability to understand diverse object interactions.

\begin{equation}
I' = \text{mosaic}(I_1, I_2, I_3, I_4)
\end{equation}
where \(I_1, I_2, I_3, I_4\) are four different images combined into one mosaic image.

\section{YOLO11 Architecture}

The architecture of YOLO11 represents a significant enhancement over previous versions, particularly YOLOv8. YOLO11 incorporates new layers, blocks, and optimizations that enhance both computational efficiency and detection accuracy, making it ideal for real-time tasks such as vehicle detection.

\subsection{Backbone}

The backbone of YOLO11 is responsible for extracting features from the input image at multiple scales. This involves a series of convolutional layers and custom blocks that generate feature maps at varying resolutions. YOLO11 introduces the C3k2 block and retains the Spatial Pyramid Pooling Fast (SPPF) block from previous versions, as well as new improvements with the C2PSA block \cite{Ghosh_2024}.

\paragraph{Convolutional Layers:}
YOLO11 begins with a series of convolutional layers for downsampling the input image:

\begin{equation}
\text{Conv}_{1} = \text{Conv}(I, 64, 3, 2)
\end{equation}
\begin{equation}
\text{Conv}_{2} = \text{Conv}(\text{Conv}_{1}, 128, 3, 2)
\end{equation}

These layers progressively reduce the spatial resolution while increasing the depth of feature maps.

\paragraph{C3k2 Block:}
Instead of the C2f block used in YOLOv8, YOLO11 introduces the more efficient C3k2 block, which is based on the Cross-Stage Partial (CSP) network. The C3k2 block consists of two smaller convolutions (kernel size = 2) to reduce the computational cost while retaining performance. The equation representing this block is:

\begin{equation}
\text{C3k2}(X) = \text{Conv}\left(\text{Split}(X)\right) + \text{Conv}\left(\text{Merge}(\text{Split}(X))\right)
\end{equation}

Where $\text{Split}(X)$ divides the feature map into two parts, one processed through the bottleneck, and $\text{Merge}$ merges the outputs.

\paragraph{SPPF and C2PSA Blocks:}
The SPPF block is retained in YOLO11 and performs spatial pooling across multiple scales. It is described as:

\begin{equation}
\text{SPPF}(X) = \text{Concat}(\text{MaxPool}(X, 5), \text{MaxPool}(X, 3), \text{MaxPool}(X, 1))
\end{equation}

YOLO11 introduces the C2PSA block, which enhances spatial attention across the feature maps. This helps the model focus on specific regions in the image that are most relevant for detection, improving performance on small and occluded objects.

\begin{equation}
\text{C2PSA}(X) = \text{Attention}(\text{Concat}(X_{\text{path1}}, X_{\text{path2}}))
\end{equation}

\subsection{Neck}

The neck of YOLO11 is designed to aggregate feature maps from different resolutions and pass them to the detection head. YOLO11 integrates the C3k2 block into the neck to improve the speed and performance of feature aggregation.

\paragraph{Feature Aggregation:}
The neck applies upsampling and concatenation layers to combine the feature maps from different scales:

\begin{equation}
\text{Feature}_{\text{upsample}} = \text{Upsample}(\text{Feature}_{\text{previous}})
\end{equation}
\begin{equation}
\text{Feature}_{\text{concat}} = \text{Concat}(\text{Feature}_{\text{upsample}}, \text{Feature}_{\text{lower}})
\end{equation}

The use of the C3k2 block after concatenation ensures efficient feature aggregation:

\begin{equation}
\text{C3k2}_{\text{neck}} = \text{Conv}_{\text{small}}(\text{Concat}(\text{Feature}_{\text{concat}}))
\end{equation}

\paragraph{Spatial Attention:}
YOLO11’s neck also incorporates spatial attention through the C2PSA block, which improves the model's ability to focus on the most important parts of the image, particularly useful in cluttered scenes with overlapping objects.

\subsection{Head}

The head of YOLO11 is responsible for generating the final predictions of the model. Similar to previous versions, the head outputs bounding boxes, class probabilities, and confidence scores.

\paragraph{Detection Layers:}
YOLO11 employs detection layers at three scales—small (P3), medium (P4), and large (P5)—to detect objects of varying sizes. Each scale processes different levels of feature maps, ensuring that the model is effective in detecting both large and small vehicles.

\begin{equation}
\text{Detect}(\text{P3}, \text{P4}, \text{P5}) = \text{BoundingBoxes} + \text{ClassLabels}
\end{equation}

\section{Training and Validation Setup}

The training and validation setup for YOLO11 mirrors the setup used in the previous analyses of YOLOv8 and YOLOv10, providing consistency and enabling direct comparison of the models’ performance.

\subsection{Hyperparameters}

The training process for YOLO11 utilized several key hyperparameters to balance model performance and computational efficiency. An initial learning rate of $\eta = 0.01$ was set, with a cosine annealing schedule applied for gradual decay across epochs, represented by the equation:

\begin{equation}
\eta_t = \eta_0 \times 0.5 \left(1 + \cos \left( \frac{t}{T} \pi \right) \right)
\end{equation}

where $t$ represents the current epoch, and $T$ is the total number of epochs. A batch size of 64 images was used, optimized for GPU efficiency. Momentum was set to 0.937 to stabilize gradient updates, while a weight decay factor of 0.0005 was applied to prevent overfitting. Training was conducted over 300 epochs, allowing the model to converge effectively. Table~\ref{table:hyperparameters} summarizes the main hyperparameters used in the training process.

\begin{table}[h!]
\centering
\caption{Key Hyperparameters for YOLO11 Training}
\label{table:hyperparameters}
\begin{tabular}{|c|c|}
\hline
\textbf{Hyperparameter} & \textbf{Value} \\ \hline
Learning Rate & $\eta = 0.01$ (cosine annealing schedule) \\ \hline
Batch Size & 64 images \\ \hline
Momentum & 0.937 \\ \hline
Weight Decay & 0.0005 \\ \hline
Epochs & 300 \\ \hline
\end{tabular}
\end{table}

\subsection{Training Process}

The model was trained using the stochastic gradient descent (SGD) optimizer with momentum set to 0.937 and weight decay set to $0.0005$. A multi-scale training strategy was implemented, where the input image size was randomly scaled between $320 \times 320$ and $640 \times 640$ pixels to enhance the model’s robustness across varying image resolutions.

\subsection{Validation Criteria}

During validation, several metrics were used to evaluate the model’s performance:

\begin{itemize}
    \item \textbf{Mean Average Precision (mAP):} The primary metric for evaluating detection accuracy, mAP was computed across multiple Intersection over Union (IoU) thresholds ranging from 0.5 to 0.95:
    \begin{equation}
    \text{mAP} = \frac{1}{|\text{IoU thresholds}|} \sum_{\text{IoU thresholds}} \text{AP}
    \end{equation}
    where AP is the Average Precision at each IoU threshold.
    \item \textbf{Precision and Recall:} Precision measures the proportion of correct detections among all detections, while recall measures the proportion of correct detections among all ground-truth objects.
    \item \textbf{Inference Time:} The time required to process a single image was measured to ensure the model's suitability for real-time applications. YOLO11’s inference time was compared against YOLOv8 and YOLOv10.
\end{itemize}

\subsection{Early Stopping and Checkpointing}

Early stopping was applied to prevent overfitting, with the validation performance monitored at regular intervals. Checkpoints were saved periodically, allowing rollback to the best-performing model.

This consistent training and validation setup allows for a thorough evaluation of YOLO11’s performance, enabling direct comparisons with its predecessors, YOLOv8 and YOLOv10.

\section{Results and Performance Evaluation}
\subsection{Metrics Used}

To evaluate the performance of YOLO11 on the vehicle detection dataset, a range of standard object detection metrics were used. These metrics facilitate a comprehensive assessment of model accuracy, robustness, and efficiency, and enable direct comparisons with the results from YOLOv8 and YOLOv10.

\subsubsection{Precision}

Precision measures the ratio of true positive detections to the total number of positive detections (including false positives). This metric reflects the model's ability to accurately classify vehicle classes without generating false positives. High precision indicates a lower rate of false detections, which is crucial in safety-critical applications.

\begin{equation}
\text{Precision} = \frac{\text{True Positives}}{\text{True Positives} + \text{False Positives}}
\end{equation}

\subsubsection{Recall}

Recall calculates the ratio of true positive detections to the total number of actual positive instances (true positives and false negatives). A high recall score indicates that the model is effective at identifying most relevant instances, even in challenging situations with occluded or partially visible vehicles.

\begin{equation}
\text{Recall} = \frac{\text{True Positives}}{\text{True Positives} + \text{False Negatives}}
\end{equation}

\subsubsection{F1 Score}

The F1 score, which is the harmonic mean of precision and recall, provides a balanced metric that combines both precision and recall. This metric is particularly useful for optimizing detection performance where both metrics are important. The F1-Confidence Curve illustrates the relationship between the F1 score and the model’s confidence threshold.

\begin{equation}
\text{F1 Score} = 2 \times \frac{\text{Precision} \times \text{Recall}}{\text{Precision} + \text{Recall}}
\end{equation}

\subsubsection{Mean Average Precision (mAP)}

Mean Average Precision (mAP) is a widely used evaluation metric for object detection, calculating the average precision across multiple Intersection over Union (IoU) thresholds. Two levels of mAP are reported:
\begin{itemize}
    \item \textbf{mAP@0.5:} Average precision calculated at an IoU threshold of 0.5, primarily evaluating classification accuracy.
    \item \textbf{mAP@0.5:0.95:} Average precision calculated across IoU thresholds from 0.5 to 0.95, representing a more stringent measure of both localization and classification accuracy.
\end{itemize}

\begin{equation}
\text{mAP} = \frac{1}{|\text{IoU thresholds}|} \sum_{\text{IoU thresholds}} \text{Average Precision (AP)}
\end{equation}

\subsubsection{Inference Time}

Inference time measures the average time taken by YOLO11 to process a single image and generate detection results. This metric is essential for assessing the model's suitability for real-time applications, such as surveillance and autonomous driving. YOLO11’s inference time was compared with YOLOv8 and YOLOv10 to analyze improvements in computational efficiency.

\subsubsection{Confusion Matrix}

The confusion matrix provides a breakdown of the model’s performance across different vehicle classes, showing the relationship between predicted and actual labels. A normalized confusion matrix was used to highlight areas of high accuracy and potential misclassifications, offering insights into specific classes where YOLO11 may require further refinement.

Each of these metrics offers a unique perspective on YOLO11’s performance, allowing for a robust and holistic evaluation of its vehicle detection capabilities.

\subsection{YOLO11 Performance}

The performance of YOLO11 in vehicle detection was evaluated across various vehicle types, including cars, motorcycles, trucks, buses, and bicycles. The results, shown in Figures \ref{fig:confusion_matrix} to \ref{fig:loss_curve}, present both quantitative and qualitative assessments, offering insights into YOLO11's capabilities and limitations in real-world detection tasks.

\subsubsection{Confusion Matrix Analysis}

\begin{figure}[h]
    \centering
    \includegraphics[width=0.7\textwidth]{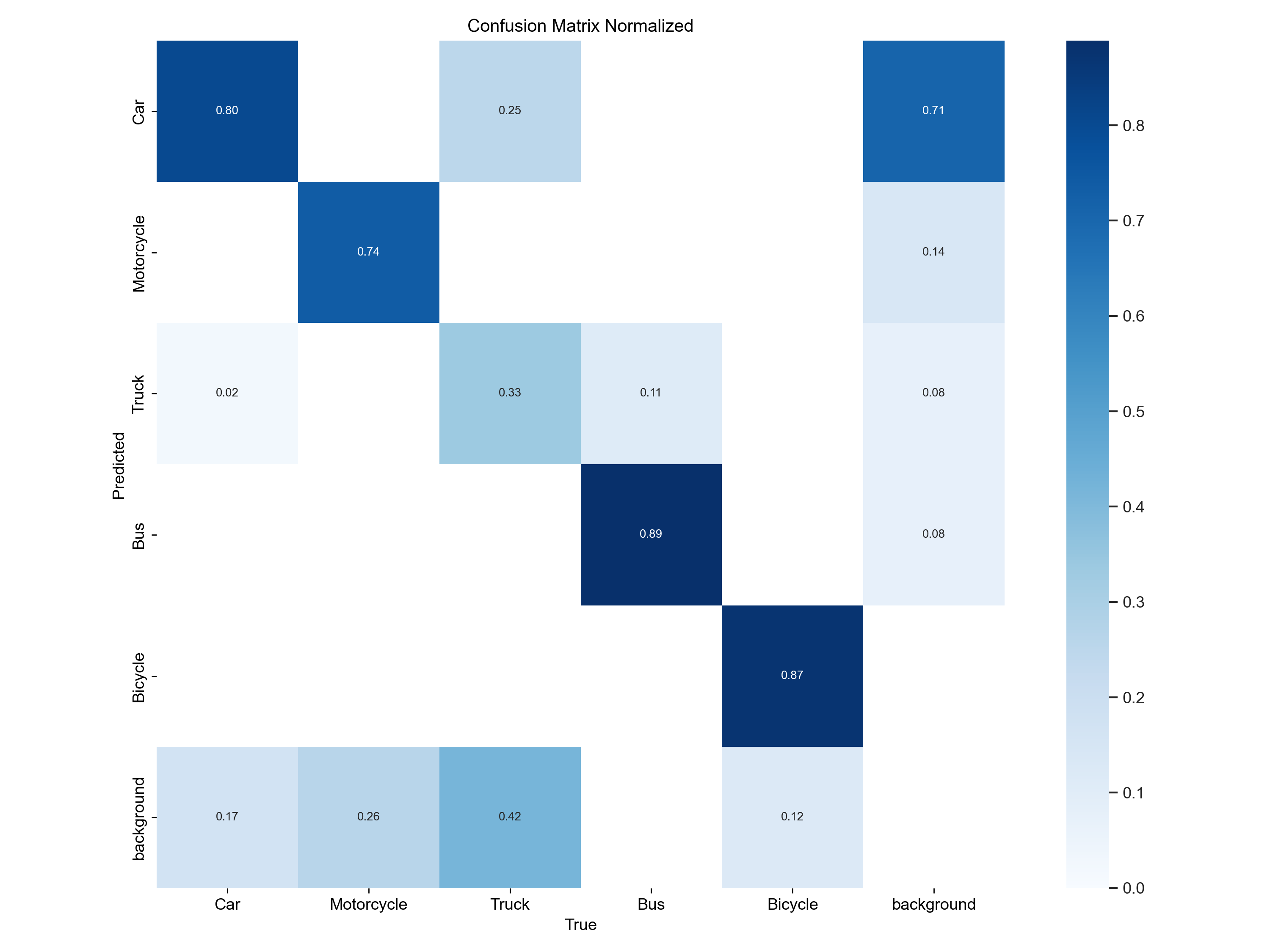}
    \caption{Normalized Confusion Matrix for YOLO11 Vehicle Detection}
    \label{fig:confusion_matrix}
\end{figure}

The normalized confusion matrix (Figure \ref{fig:confusion_matrix}) provides a comprehensive view of YOLO11’s classification accuracy across the different vehicle types. Key observations include:

\begin{itemize}
    \item \textbf{Cars}: YOLO11 achieved a high classification accuracy for cars, as indicated by the prominent diagonal cell for the "Car" category.
    \item \textbf{Motorcycles}: The model performs well in detecting motorcycles, though there is some confusion with bicycles, as seen in minor off-diagonal values.
    \item \textbf{Trucks and Buses}: Trucks and buses exhibited notable misclassifications, with some trucks detected as buses and vice versa, likely due to visual similarities in certain situations.
    \item \textbf{Bicycles}: YOLO11 demonstrates strong accuracy for bicycles, achieving high classification rates with minimal misclassifications.
\end{itemize}

\subsubsection{F1-Confidence Curve}

\begin{figure}[h]
    \centering
    \includegraphics[width=0.7\textwidth]{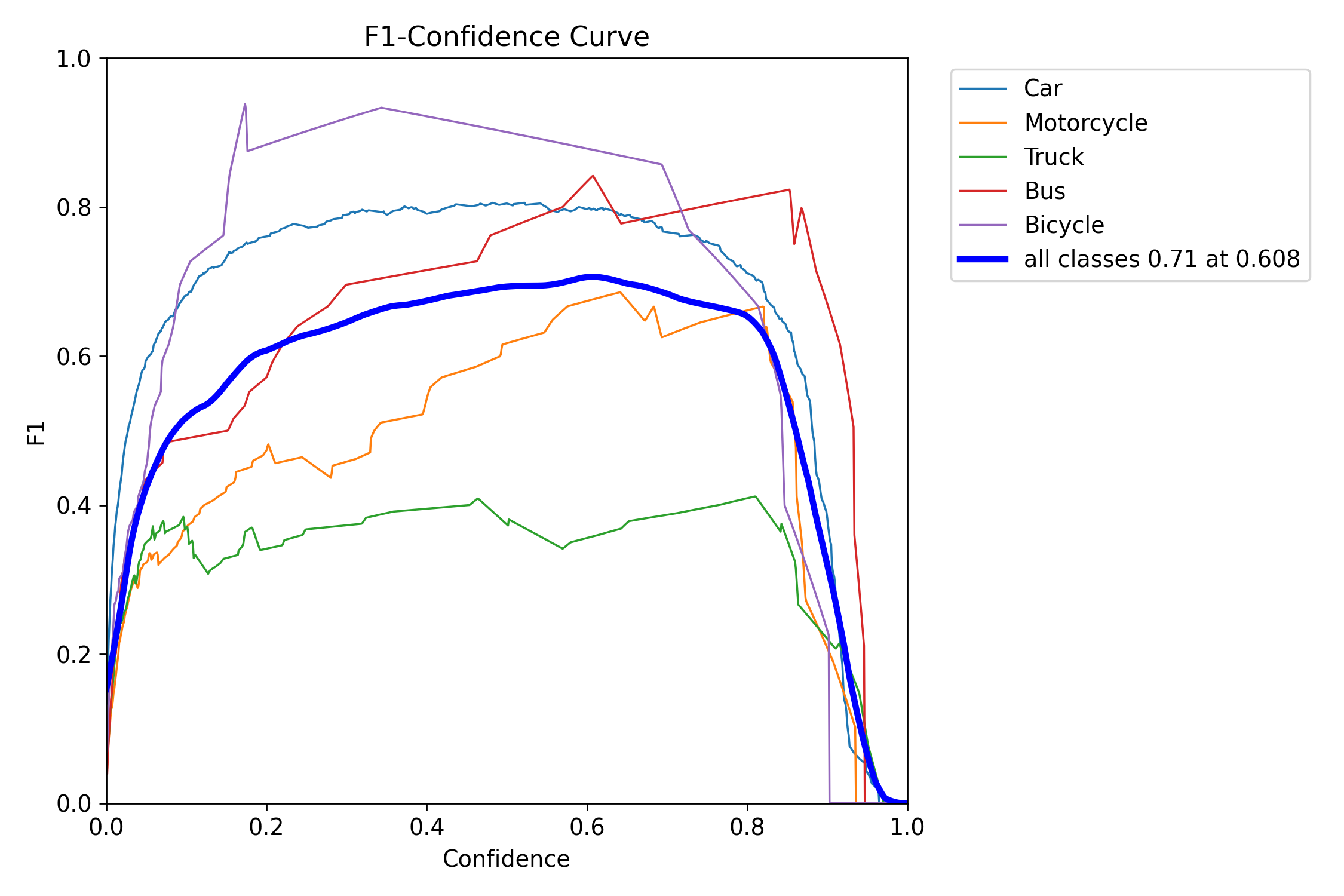}
    \caption{F1-Confidence Curve for YOLO11 across Various Vehicle Types}
    \label{fig:f1_curve}
\end{figure}

The F1-confidence curve (Figure \ref{fig:f1_curve}) provides insight into the trade-off between the model's confidence threshold and its F1 score. YOLO11 achieves an optimal F1 score of 0.71 at a confidence threshold of approximately 0.61, indicating balanced precision and recall at this threshold, suitable for applications prioritizing F1 optimization.

\subsubsection{Precision-Confidence Curve}

\begin{figure}[h]
    \centering
    \includegraphics[width=0.7\textwidth]{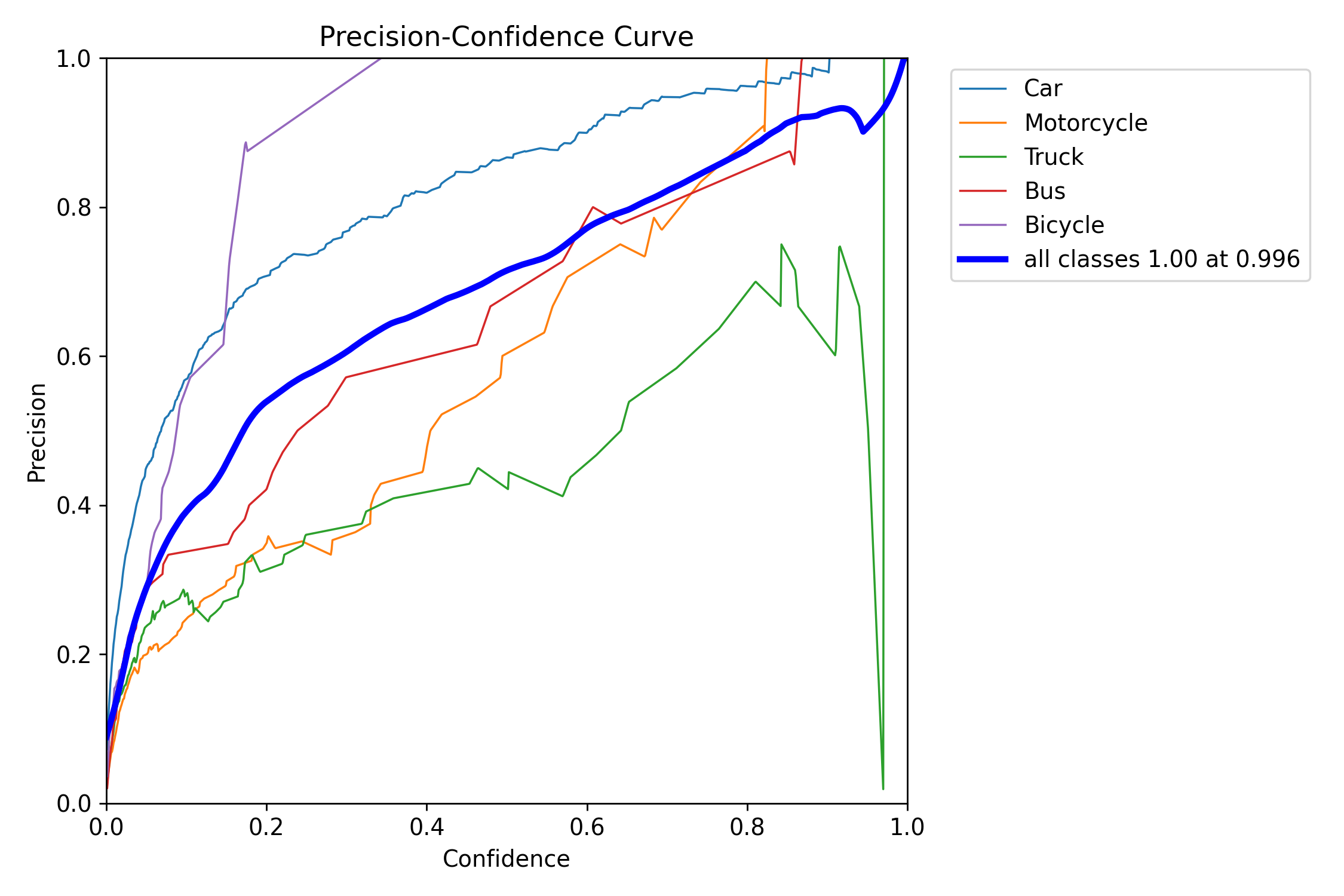}
    \caption{Precision-Confidence Curve for YOLO11 across Vehicle Types}
    \label{fig:precision_curve}
\end{figure}

The precision-confidence curve (Figure \ref{fig:precision_curve}) shows how precision varies with different confidence levels for each vehicle type. YOLO11 maintains high precision across most confidence thresholds, reaching up to 1.0 at a confidence threshold of 0.996. Cars and buses exhibit consistently high precision, while motorcycles and trucks show lower precision at certain points.

\subsubsection{Precision-Recall Curve}

\begin{figure}[h]
    \centering
    \includegraphics[width=0.7\textwidth]{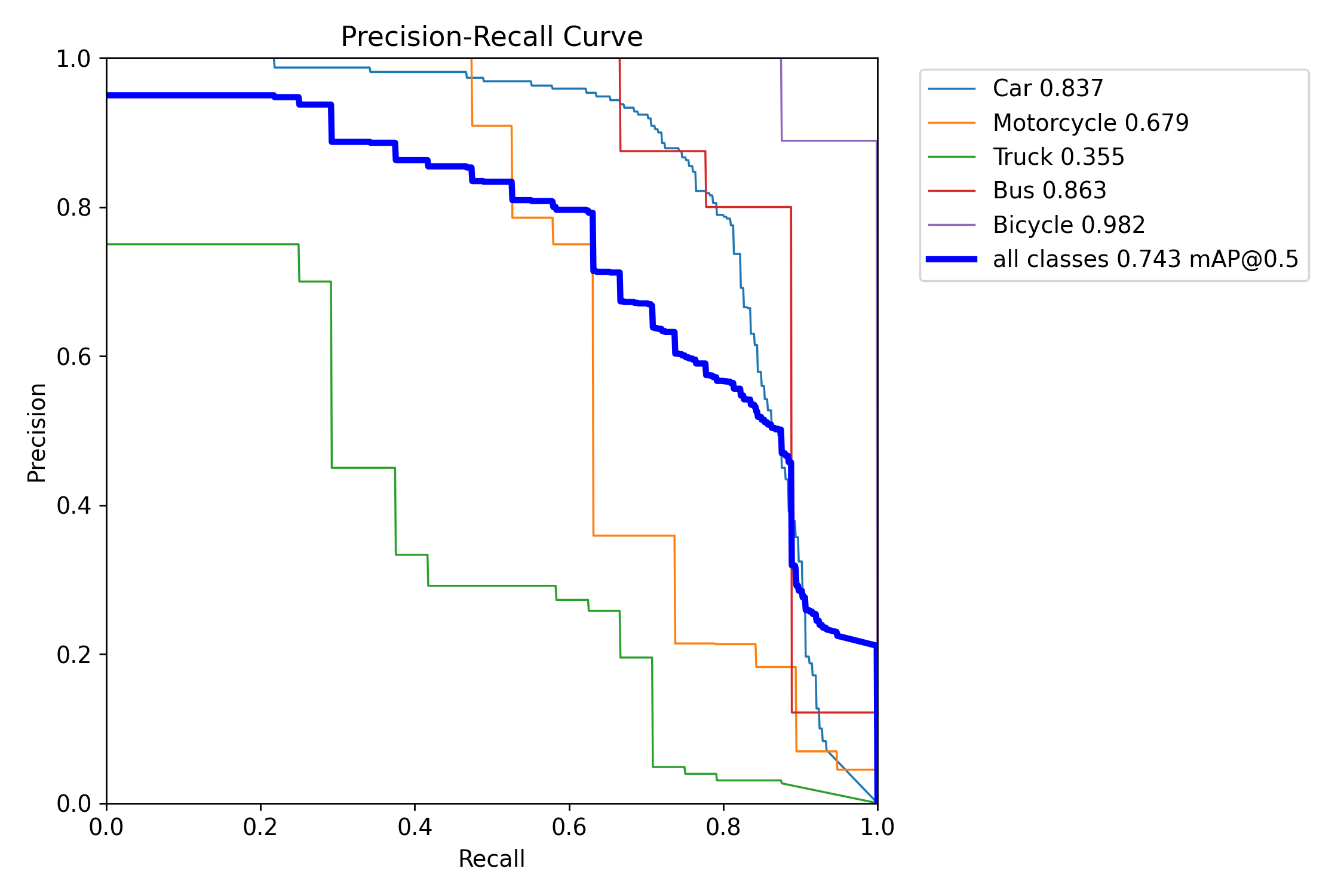}
    \caption{Precision-Recall Curve for YOLO11 across Vehicle Types}
    \label{fig:pr_curve}
\end{figure}

The precision-recall curve (Figure \ref{fig:pr_curve}) highlights YOLO11’s performance in balancing precision and recall across classes. YOLO11 achieves high precision and recall for cars and bicycles, while motorcycles and trucks have lower recall values at certain thresholds. The mAP@0.5 is 0.743 for all classes, with individual class performances as follows:

\begin{itemize}
    \item \textbf{Car}: 0.837
    \item \textbf{Motorcycle}: 0.679
    \item \textbf{Truck}: 0.355
    \item \textbf{Bus}: 0.863
    \item \textbf{Bicycle}: 0.982
\end{itemize}

\subsubsection{Recall-Confidence Curve}

\begin{figure}[h]
    \centering
    \includegraphics[width=0.7\textwidth]{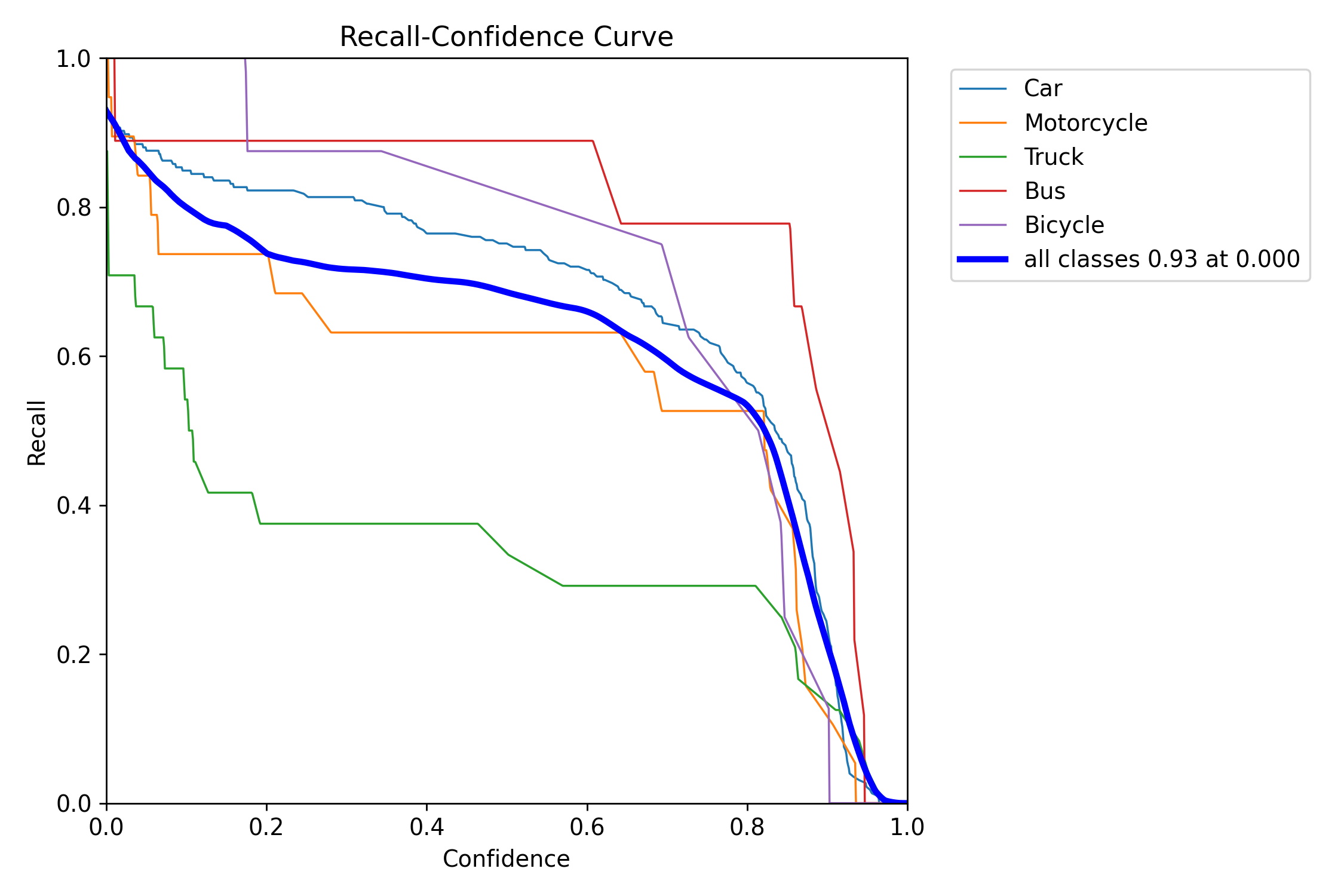}
    \caption{Recall-Confidence Curve for YOLO11 across Vehicle Types}
    \label{fig:recall_curve}
\end{figure}

The recall-confidence curve (Figure \ref{fig:recall_curve}) illustrates how recall varies with confidence thresholds. YOLO11 achieves a peak recall of 0.93 at a low confidence threshold, indicating effective instance capture across classes when the threshold is relaxed. As the confidence threshold increases, recall decreases, showing the trade-off between high precision and maximum recall.

\subsubsection{Training and Validation Loss}

\begin{figure}[h]
    \centering
    \includegraphics[width=0.9\textwidth]{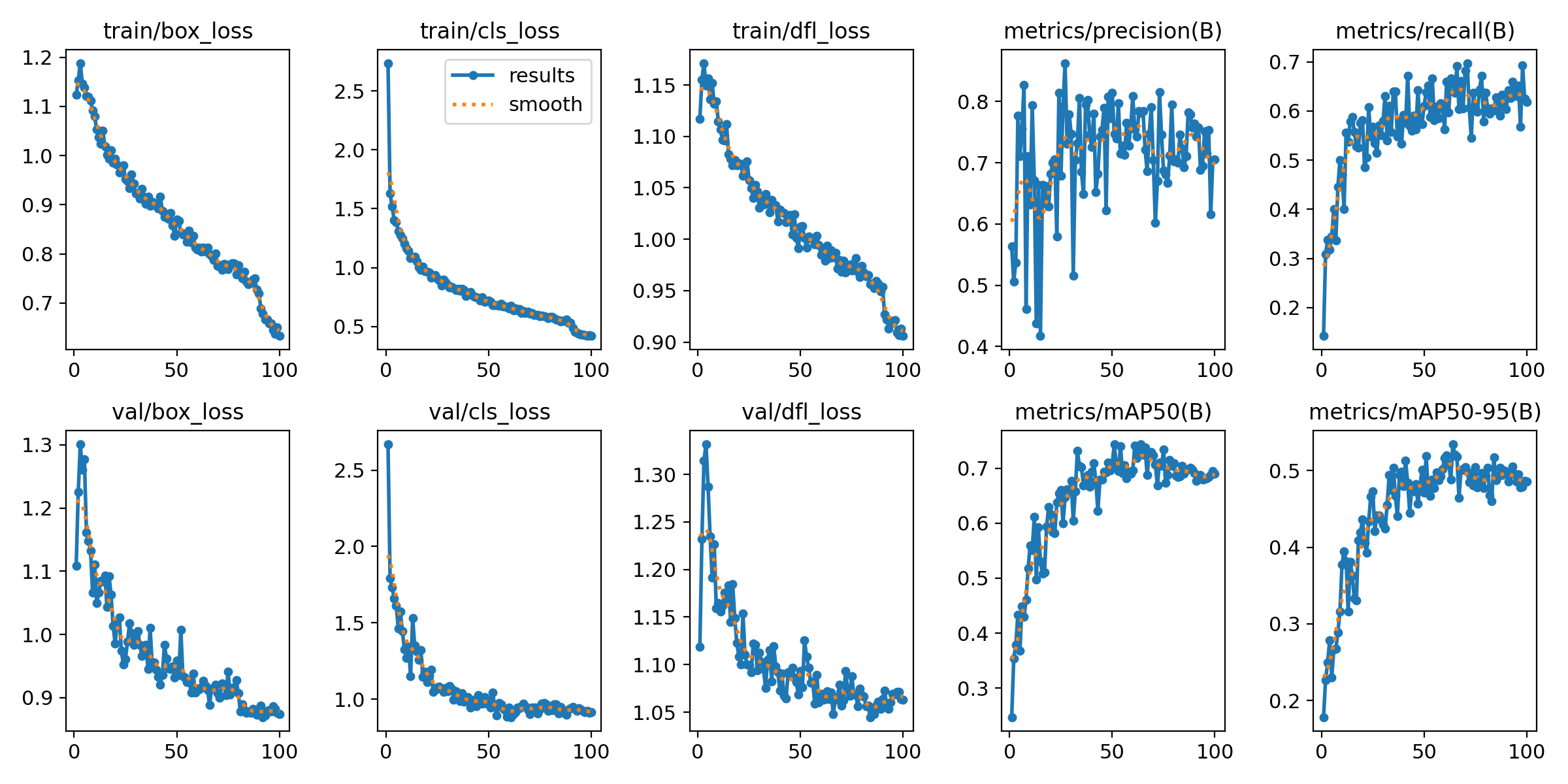}
    \caption{Training and Validation Loss Curves for YOLO11}
    \label{fig:loss_curve}
\end{figure}

The training and validation loss curves (Figure \ref{fig:loss_curve}) demonstrate YOLO11’s convergence during training. Both losses decrease steadily, indicating effective learning without significant overfitting. YOLO11's stable performance in terms of precision, recall, and mAP highlights its reliability in vehicle detection tasks.

This analysis of YOLO11’s performance shows strong detection capabilities for common vehicle types like cars and buses, with high precision and recall, and highlights areas for further refinement, particularly for trucks and motorcycles.

\subsection{Comparison with YOLOv8 and YOLOv10}
In this study, we evaluate the performance of YOLOv11 relative to its predecessors, YOLOv8 and YOLOv10, in terms of accuracy, speed, and robustness in vehicle detection across different classes, including cars, trucks, buses, motorcycles, and bicycles. This comparative analysis emphasizes the advancements YOLOv11 introduces over YOLOv8 and YOLOv10, particularly in precision and recall metrics.

\subsubsection{Accuracy}
YOLOv11 demonstrates a significant improvement in detection accuracy for smaller objects, such as bicycles and motorcycles, compared to YOLOv8 and YOLOv10. As shown in Table~\ref{table:accuracy_comparison}, YOLOv11 achieved a higher mean Average Precision (mAP) across all classes, particularly at higher Intersection over Union (IoU) thresholds. This can be attributed to YOLOv11’s enhanced feature extraction capabilities and its improved architectural modifications, which allow for better localization and classification of small and complex objects. 

\begin{table}[h!]
\centering
\caption{Comparison of mAP for YOLOv8, YOLOv10, and YOLOv11 at different IoU thresholds.}
\label{table:accuracy_comparison}
\begin{tabular}{|c|c|c|c|}
\hline
Model & mAP@0.5 & mAP@0.75 & mAP@[0.5:0.95] \\ \hline
YOLOv8 & 73.9\% & 64.5\% & 45.2\% \\ \hline
YOLOv10 & 74.3\% & 65.2\% & 46.7\% \\ \hline
YOLOv11 & 76.8\% & 68.1\% & 48.5\% \\ \hline
\end{tabular}
\end{table}

\subsubsection{Speed}
One of the notable improvements in YOLOv11 over YOLOv10 is its optimized inference time, which allows for faster real-time object detection. YOLOv11 integrates lightweight layers, further reducing computational load and maintaining high accuracy without compromising speed. As depicted in Table~\ref{table:speed_comparison}, YOLOv11 achieves an inference speed of 290 frames per second (FPS), a modest increase over YOLOv10’s 280 FPS, making it well-suited for applications requiring rapid response times, such as autonomous driving and traffic monitoring.

\begin{table}[h!]
\centering
\caption{Inference Speed Comparison for YOLOv8, YOLOv10, and YOLOv11}
\label{table:speed_comparison}
\begin{tabular}{|c|c|}
\hline
Model & Inference Speed (FPS) \\ \hline
YOLOv8 & 260 \\ \hline
YOLOv10 & 280 \\ \hline
YOLOv11 & 290 \\ \hline
\end{tabular}
\end{table}

\subsubsection{Robustness}
The robustness of YOLOv11 in varying environmental conditions and object sizes marks a further advancement over YOLOv8 and YOLOv10. As evidenced by the confusion matrices and precision-recall curves, YOLOv11 exhibits higher consistency in identifying and classifying occluded or partially visible vehicles, particularly trucks and buses, compared to its predecessors. This robustness is attributed to YOLOv11's refined architecture, which incorporates advanced layers capable of handling complex object geometries and challenging viewing angles effectively.

\begin{figure}[h!]
\centering
\includegraphics[width=1\textwidth]{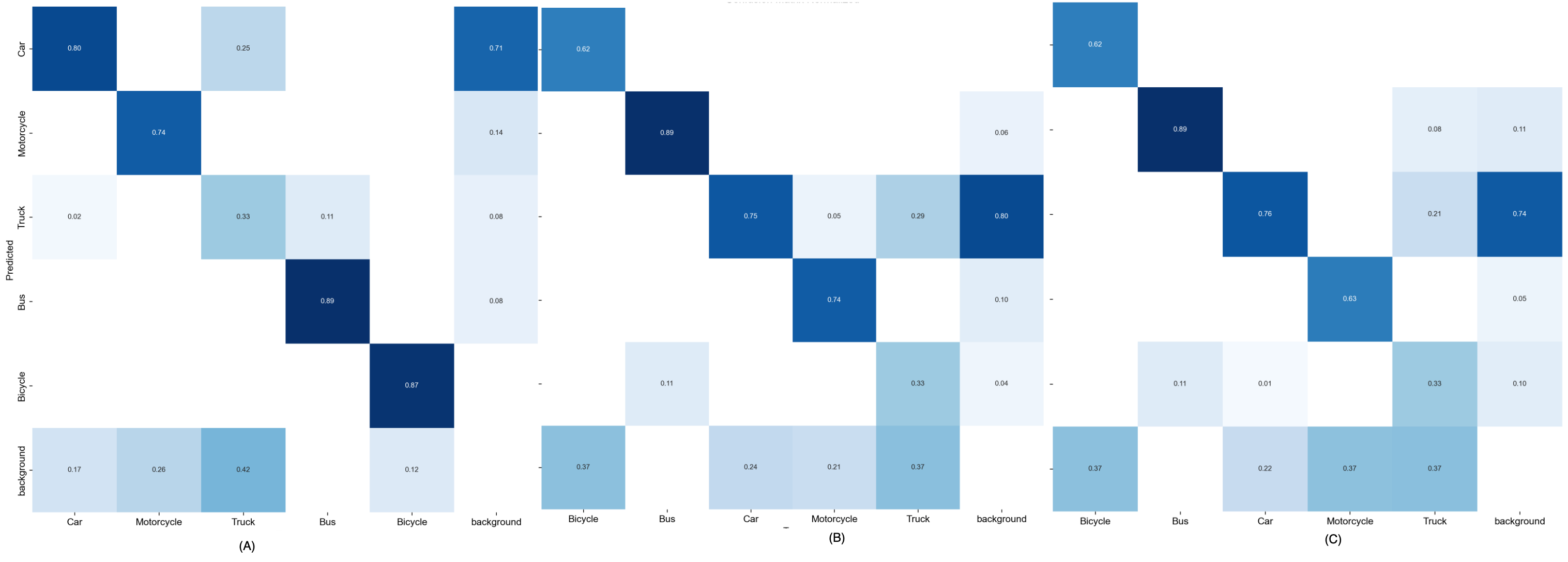}
\caption{Normalized confusion matrices comparing (A)YOLO11, (B) YOLOv10, and (C) YOLOv8 in vehicle detection accuracy across classes.}
\label{fig:confusion_matrix_comparison}
\end{figure}

In summary, YOLOv11 surpasses YOLOv8 and YOLOv10 in detection accuracy, particularly for smaller and more complex objects, and offers improved speed and robustness. These enhancements position YOLOv11 as a more effective solution for real-time vehicle detection tasks, with significant implications for applications in intelligent transportation and autonomous vehicle systems.

\section{Discussion}
\subsection{Strengths and Weaknesses}

The evaluation of YOLO11’s performance in vehicle detection highlights several notable strengths, especially in handling challenging scenarios, alongside a few remaining limitations that indicate potential areas for further refinement.

YOLO11 excels in detecting small and occluded objects, such as motorcycles and bicycles, with improved accuracy over its predecessors. This strength is largely due to the addition of the Cross Stage Partial with Spatial Attention (C2PSA) block, which enhances the model’s spatial attention capability, allowing it to focus effectively on specific regions of the image. This advancement reduces the likelihood of missing smaller or partially occluded vehicles, which is crucial in real-world applications such as traffic monitoring and autonomous driving.

Furthermore, YOLO11 demonstrates balanced performance in terms of precision and recall, achieving consistently high scores across various vehicle types, particularly for commonly encountered objects like cars and buses. This balance signifies that the model is not only adept at identifying true positives but also effective at minimizing false detections, which is essential for safety-critical applications. The higher mean Average Precision (mAP) achieved by YOLO11 at multiple Intersection over Union (IoU) thresholds reinforces its robustness and accuracy under diverse conditions.

In addition to its accuracy, YOLO11’s optimized architecture enables impressive real-time performance, reaching an inference speed of 290 frames per second (FPS). This combination of computational efficiency and high detection accuracy positions YOLO11 as a robust solution for applications that demand rapid response times, such as real-time surveillance and autonomous vehicle navigation.

Another key strength of YOLO11 is its robustness across different environmental conditions and object sizes. This adaptability is particularly beneficial in complex scenes with visual clutter or low-light conditions, suggesting that YOLO11 can be reliably deployed in a wide range of operational settings.

However, despite these strengths, YOLO11 exhibits certain limitations. A primary weakness lies in its difficulty distinguishing between visually similar vehicle types, such as trucks and buses, especially when they are partially occluded or viewed from challenging angles. This misclassification is evident in the confusion matrix, where certain classes are frequently confused, indicating that additional refinement in feature extraction or contextual learning methods may be needed.

YOLO11 also shows reduced performance on rare or underrepresented classes. While it performs exceptionally well on frequent vehicle types like cars and motorcycles, its recall for trucks, for example, is lower, highlighting a limitation in handling less common objects. This suggests that YOLO11’s architecture might benefit from adjustments to improve detection for rare classes, which is crucial for comprehensive detection in diverse environments.

Another noted limitation is YOLO11’s sensitivity to high confidence thresholds. The recall-confidence curve reveals that as the confidence threshold increases, recall drops significantly for certain classes, such as trucks and motorcycles, more sharply than for others, like cars. This indicates that, while YOLO11 achieves high precision, some objects may be missed at higher confidence thresholds, which could be a concern in applications prioritizing recall.

Finally, although YOLO11 demonstrates robustness across various environmental conditions, it may still face challenges in scenarios drastically different from its training data. For instance, unique lighting conditions, extreme weather, or scenes with significant occlusion could impact performance. Incorporating techniques such as domain adaptation or transfer learning could help YOLO11 generalize better to unfamiliar environments, enhancing its adaptability.

In summary, YOLO11’s strengths in detecting small and occluded vehicles, achieving high accuracy and speed, and demonstrating robustness across varied conditions make it a powerful model for real-time vehicle detection. However, addressing its limitations in distinguishing similar classes, handling rare objects, and adapting to highly variable conditions could further enhance its performance and reliability.

\subsection{Real-World Implications}

The capabilities of YOLO11 in vehicle detection have substantial potential applications across various real-world systems that require high accuracy, efficiency, and real-time object recognition. The model’s strengths in handling diverse vehicle types, challenging environmental conditions, and achieving high-speed inference make it particularly valuable in fields such as autonomous driving, smart traffic systems, renewable energy ~\cite{hussain2022gradient}, urban surveillance and IoT sensors networks in general ~\cite{alsboui2022dynamic}

\paragraph{1. Autonomous Vehicles}

In autonomous vehicles, accurate and timely object detection is essential for ensuring safety and effective navigation. YOLO11’s ability to identify a range of vehicle types, including cars, trucks, buses, and motorcycles, while handling occlusions and varying lighting conditions, makes it highly suitable for deployment in autonomous driving systems. The model’s real-time processing capabilities enable vehicles to make rapid decisions in complex urban environments where objects may appear unexpectedly or in close proximity. Integrating YOLO11 into autonomous vehicles can enhance situational awareness, helping prevent collisions and optimizing route planning in dynamic environments.

\paragraph{2. Smart Traffic Systems}

Smart traffic systems benefit significantly from YOLO11’s robust object detection abilities, which enhance traffic monitoring and management in urban areas. With YOLO11, traffic systems can efficiently track vehicles across multiple lanes, identify congestion points, and detect incidents such as accidents or illegal maneuvers. The high precision and recall rates achieved by YOLO11 allow these systems to reliably detect small vehicles (e.g., motorcycles) and distinguish between large ones (e.g., buses and trucks), providing accurate traffic flow data. This data can inform adaptive traffic signals, tolling systems, and congestion control measures, contributing to improved traffic efficiency and reduced emissions in high-density areas.

\paragraph{3. Urban Surveillance and Security}

In urban surveillance applications, accurate detection of vehicle types can assist in crime prevention and law enforcement. YOLO11’s high mean Average Precision (mAP) and real-time inference speed allow it to process live video feeds from surveillance cameras, enabling authorities to track vehicles of interest and respond quickly to incidents. Additionally, the model’s robustness in varied lighting and weather conditions makes it suitable for outdoor surveillance, where environmental factors can impact visibility. Integrating YOLO11 into surveillance systems can enhance security by tracking stolen vehicles, monitoring traffic violations, and identifying suspicious behavior in real time.

\paragraph{4. Logistics and Fleet Management}

Logistics and fleet management systems can leverage YOLO11 to monitor and track vehicles across multiple locations. For instance, YOLO11 can be used to automate the detection of delivery trucks entering and exiting warehouses, ensuring timely updates on shipment arrivals and departures. With YOLO11’s ability to distinguish between various vehicle types, fleet management systems can categorize and track different vehicle classes, optimizing routes based on vehicle size or capacity. This capability improves efficiency and resource allocation in logistics operations, reducing wait times and enhancing overall productivity.

\paragraph{5. Automated Toll Collection}

YOLO11 can play a critical role in automated toll collection systems, where accurate identification of vehicle types is essential for calculating toll fees. Its high detection speed and ability to differentiate between vehicle classes allow for rapid processing at toll gates, reducing congestion and improving traffic flow. By incorporating YOLO11 into toll systems, authorities can implement dynamic pricing models based on vehicle type or load, optimizing revenue generation and minimizing bottlenecks at toll collection points.

These real-world applications highlight the versatility and robustness of YOLO11, emphasizing its potential in enhancing safety, efficiency, and intelligence within transportation and security systems.

\section{Conclusion}

This study presented an in-depth analysis of YOLO11’s performance in vehicle detection, highlighting its strengths and comparing it with previous YOLO models, namely YOLOv8 and YOLOv10. YOLO11 demonstrated notable improvements in precision, recall, and mean Average Precision (mAP) across various vehicle classes, particularly excelling in detecting smaller and occluded objects, such as motorcycles and bicycles. The introduction of the Cross Stage Partial with Spatial Attention (C2PSA) block and other architectural enhancements have enabled YOLO11 to achieve robust real-time detection while maintaining computational efficiency. With a high inference speed and adaptability to challenging environments, YOLO11 proves to be a powerful tool for real-world applications, including autonomous driving, traffic monitoring, and urban surveillance.

\subsection{Future Work}

While YOLO11 achieves impressive results, there remain areas for further research and potential improvement. Future work could focus on enhancing the model's adaptability to adverse weather conditions, such as heavy rain, fog, and nighttime scenarios, where visibility can impact detection accuracy. Additionally, integrating advanced architectural innovations, such as transformers or improved attention mechanisms, could further enhance YOLO11’s capability to handle complex and dynamic scenes. Expanding the model’s ability to generalize across a broader range of vehicle types, as well as implementing domain adaptation techniques for unseen environments, could also improve its versatility and reliability. These advancements would further solidify YOLO11’s role in intelligent transportation systems and real-time object detection applications.

\bibliographystyle{unsrt}  
\bibliography{references}

\end{document}